\documentclass[final,5p,times,twocolumn]{elsarticle}

\usepackage{amssymb}
 \usepackage{amsthm}

\biboptions{square,sort&compress}

\usepackage[cmex10]{amsmath}
\usepackage{verbatim} 
\usepackage{mathrsfs} 
\usepackage{algorithm}
\usepackage{algorithmic}
\usepackage{l3keys2e} 
\usepackage{siunitx} 
\usepackage[labelsep=period,font={footnotesize}]{caption} 
\usepackage{subfig}
\usepackage{placeins} 

\usepackage{booktabs} 
\usepackage{multirow}

\DeclareMathOperator*{\rank}{rank}

\newcommand{\eqnref}[1]{ (\ref{#1})}
\newcommand{\eqnsref}[2]{(\ref{#1}-\ref{#2})}

\newcommand{\figref}[1]{Fig. \ref{#1}}
\newcommand{\secref}[1]{Section \ref{#1}}

\newcommand{\transp}{^\mathsf{T}}

\newcommand{\invtransp}{^{-\mathsf{T}}}

\newcommand{\identity}{ \mathtt{I} }

\begin{document}

\begin{frontmatter}




\title{Internal Constraints of the Trifocal Tensor}


\author[ncsu]{Stuart B. Heinrich\corref{cor1} and Wesley E. Snyder}
\cortext[cor1]{Corresponding author}
\ead{sbheinri@ncsu.edu}

\address[ncsu]{Department of Electrical and Computer Engineering, NC State University, Box 7911, Raleigh, NC 27695-7911}

\begin{abstract}
The fundamental matrix and trifocal tensor are convenient algebraic representations of the epipolar geometry of two and three view configurations, respectively.  The estimation of these entities is central to most reconstruction algorithms, and a solid understanding of their properties and constraints is therefore very important.  The fundamental matrix has 1 internal constraint which is well understood, whereas the trifocal tensor has 8 independent algebraic constraints.  The internal tensor constraints can be represented in many ways, although there is only one minimal and sufficient set of 8 constraints known.  In this paper, we derive a second set of minimal and sufficient constraints that is simpler.  We also show how this can be used in a new parameterization of the trifocal tensor.  We hope that this increased understanding of the internal constraints may lead to improved algorithms for estimating the trifocal tensor, although the primary contribution is an improved theoretical understanding.
\end{abstract}


\end{frontmatter}

\section{Introduction}

It is well known that a reconstruction from projection constraints alone is, at best, ambiguous up to an arbitrary projective transform having 15 degrees of freedom (dof).  Each camera matrix has 11 dof, so there are $11m - 15$ dof to the projective-invariant geometry (henceforth referred to as \emph{epipolar geometry}) of $m$ unique cameras \citep[sec. 17.5]{HARTLEY2004}.

In the case of 2 views, the epipolar geometry has 7 dof and may be conveniently represented by the $3 \times 3$ \emph{fundamental matrix} $\mathbf{F}$.  The additional 2 parameters in the matrix $\mathbf{F}$ may be attributed to an ambiguous overall scale factor and a single internal constraint that $\rank \mathbf{F} = 2$.  This is usually relaxed to $\rank \mathbf{F} \leq 2$, making it equivalent to $\det \mathbf{F} = 0$, which can be written as a simple polynomial.

Similarly, the geometric relationship between three views can be conveniently represented by a $3\times3\times3$ tensor known as the \emph{trifocal tensor} and denoted by $\mathcal{T}$.  Again by the argument above, it is clear that the epipolar geometry of three views has 18 dof, although the tnesor $\mathcal{T}$ has 27 parameters.  Thus, not all tensors represent consistent epipolar geometry. One parameter may be attributed to the overall scale factor, meaning that a \emph{geometrically valid} tensor must satisfy 8 independent algebraic constraints.

The constraints on corresponding lines in three views were first derived for calibrated cameras in \citep{Spetsakis90, Weng92}.  These constraints were generalized to the uncalibrated case in \citep{SHASHUA95Algebraic}, and formulated in terms of a trifocal tensor in \citep{HARTLEY95Linear, SHASHUA95Trilinearity, TRIGGS95}.  It was shown in \citep{HARTLEY97Lines} that point constraints could also be represented using the tensor.

Estimation of the trifocal tensor has become an integral part of many projective reconstruction algorithms, for example \citep{FITZGIBBON98Automatic, NISTER01}, so these constraints are of considerable interest.  However, unlike the case with the fundamental matrix, the trifocal constraints are not so straightforward and are still not fully understood \citep{FAUGERAS95Geometry, LAVEAU96, TORR97} \citep[sec. 15.1]{HARTLEY2004}.

A set of 32 necessary and sufficient constraints were derived in \citep{FAUGERAS98Nonlinear}, and a different set of 12 necessary and sufficient constraints were derived in \citep{PAPADOPOULO98}.  In \citep{CANTERAKIS00}, the first minimal set of 8 constraints were derived.

In this paper, we derive a new class of constraints that can be taken with 5 previously known constraints to yield a much simpler set of minimal constraints.  This is a very satisfying theoretical finding and we believe it may be the last type of constraint that exists.  We also show how these new constraints can be used to parameterize the tensor, and finally discuss how a sufficient set of constraints can be formulated as fixed polynomials if desired.

\section{Definition of the Tensor}

Let us begin with a definition of the tensor as described in \citep{HARTLEY2004}.  Without loss of generality, the first projection matrix can be assumed canonical, so that the set of projection matrices for three views can be written as

\begin{align}
\mathbf{P} &= [ \identity | \mathbf{0} ] \\
\mathbf{P}' &= [ \mathbf{a}_1 \ldots \mathbf{a}_4 ] = [ \mathbf{A} | \mathbf{a}_4 ]\\
\mathbf{P}'' &= [ \mathbf{b}_1 \ldots \mathbf{b}_4 ] = [ \mathbf{B} | \mathbf{b}_4 ]\label{formP3} \text{.}
\end{align}

The tensor will be derived based on a correspondence between images of a line in 3D space.  Let the three corresponding lines in the image plane be denoted as $\mathbf{l} \leftrightarrow \mathbf{l}' \leftrightarrow \mathbf{l}''$.  The back projection of each line yields a plane,

\begin{align}
\pi &= \mathbf{P} \transp \mathbf{l} = ( \mathbf{l} \transp,0) \transp \label{plane1} \\
\pi' &= {\mathbf{P}'} \transp \mathbf{l}' =
\left[
  \begin{array}{c}
    \mathbf{A} \transp \mathbf{l}' \\
    \mathbf{a}_4 \transp \mathbf{l}' \\
  \end{array}
\right] \\
\pi'' &= {\mathbf{P}''} \transp \mathbf{l}'' =
\left[
  \begin{array}{c}
    \mathbf{B} \transp \mathbf{l}'' \\
    \mathbf{b}_4 \transp \mathbf{l}'' \\
  \end{array}
\right] \label{plane3} \text{.}
\end{align}

Because the lines were all images of a single 3D line, these back-projected planes must all intersect in a single 3D line which we write parametrically as a linear combination of two points $\mathbf{X}_1$ and $\mathbf{X}_2$,

\begin{align}
\mathbf{X}(t) = t \mathbf{X}_1 + (1-t) \mathbf{X}_2 \text{.}
\end{align}

\begin{figure}[H]
\centering
\includegraphics[width=3in]{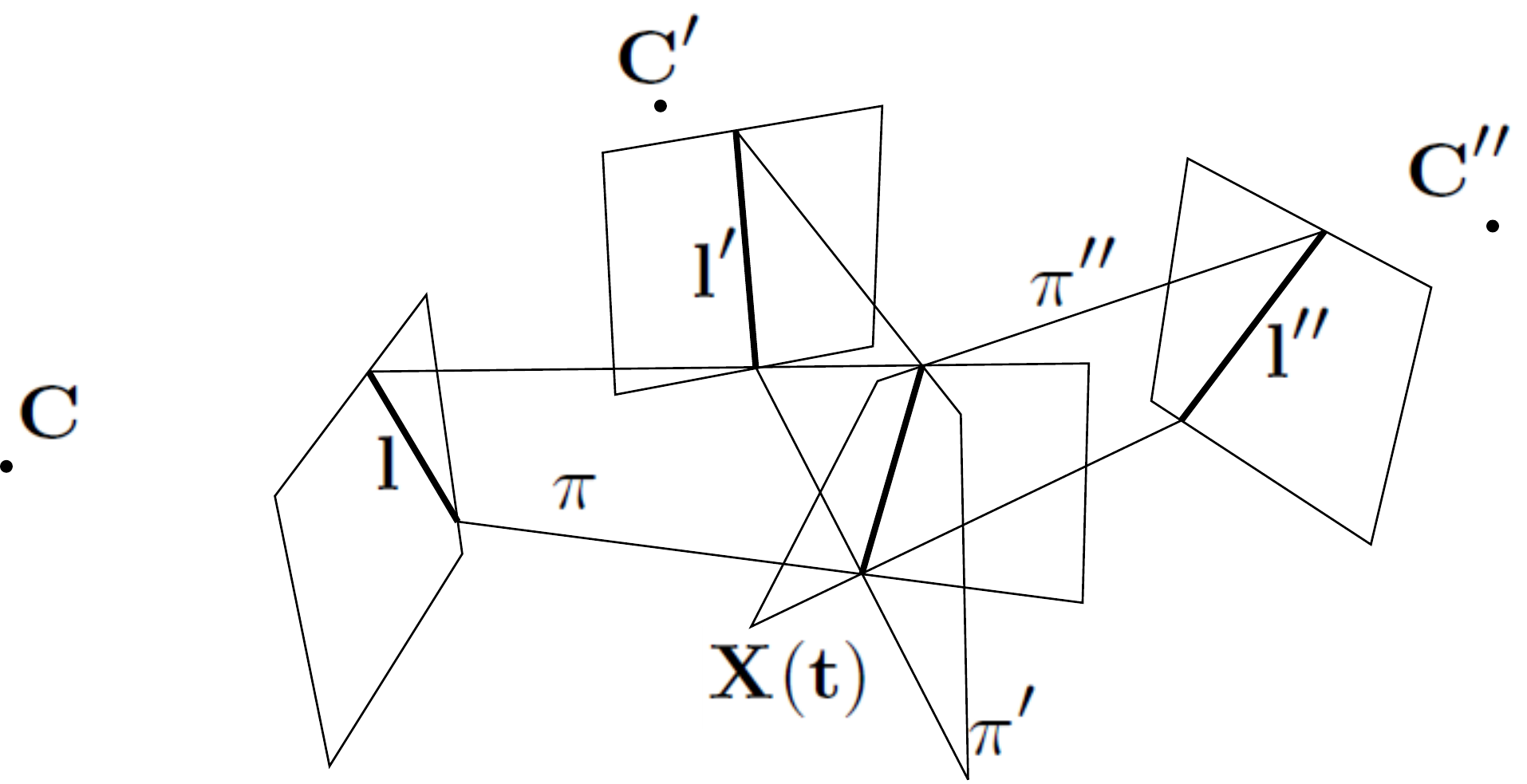}
\caption{Diagram of trifocal line constraints.  The first camera center is denoted by $\mathbf{C}$.  A parametric 3D line in space is given by $X(t)$.  This line projects onto the first image plane as $\mathbf{l}$. The line $\mathbf{l}$ back-projects to the plane $\pi$.  Notation is similar with respect to the other two views.}
\label{fig_planes}
\end{figure}

This incidence relation is diagrammed in \figref{fig_planes}.  Clearly, $\mathbf{X}(t)$ must be a point on each back-projected plane equation, so

\begin{align}
\pi \transp \mathbf{X}(t) = {\pi'} \transp \mathbf{X}(t) = {\pi''} \transp \mathbf{X}(t) = 0 \text{.}
\end{align}

If we concatenate these planes into a $4 \times 3$ matrix $\mathbf{M} = [ \pi | \pi' | \pi'' ]$, then $\mathbf{M} \transp \mathbf{X}(t) = 0$, which implies $\mathbf{M} \transp \mathbf{X}_1 = 0$ and $\mathbf{M} \transp \mathbf{X}_2 = 0$.  Thus, $\mathbf{M}$ has a 2-dimensional null space and must be rank 2 from the rank-nullity theorem.  Substituting \eqnsref{plane1}{plane3} into $\mathbf{M}$, we obtain

\begin{align}
\mathbf{M} = [ \mathbf{m}_1 | \mathbf{m}_2 | \mathbf{m}_3 ] = \left[
  \begin{array}{ccc}
    \mathbf{l} & \mathbf{A} \transp \mathbf{l}' & \mathbf{B} \transp \mathbf{l}'' \\
    0 & \mathbf{a}_4 \transp \mathbf{l}' & \mathbf{b}_4 \transp \mathbf{l}'' \\
  \end{array}
\right] \text{.}
\end{align}

Because $\mathbf{M}$ is of rank 2, it follows that the first column can be written as a linear combination of the second two columns, so $\mathbf{m}_1 = \alpha \mathbf{m}_2 + \beta \mathbf{m}_3$.  From the bottom row we obtain,

\begin{align}
0 = \alpha \mathbf{a}_4 \transp \mathbf{l}' + \beta \mathbf{b}_4 \transp \mathbf{l}'' \text{,}
\end{align}

\noindent which implies that $\alpha = k \mathbf{b}_4 \transp \mathbf{l}''$ and $\beta = -k \mathbf{a}_4 \transp \mathbf{l}'$ for some scalar $k$.  Making these substitutions back into the top half of $\mathbf{M}$ provides a homogeneous equivalence constraint between the lines,

\begin{align}
\mathbf{l} &= \mathbf{b}_4 \transp \mathbf{l}'' \mathbf{A} \transp \mathbf{l}' - \mathbf{a}_4 \transp
\mathbf{l}' \mathbf{B} \transp \mathbf{l}'' \\
&= {\mathbf{l}''} \transp \mathbf{b}_4 \mathbf{A} \transp \mathbf{l}' - {\mathbf{l}'}\transp \mathbf{a}_4 \mathbf{B} \transp \mathbf{l}'' \label{orig_constr} \text{.}
\end{align}

Introducing the notation $\mathbf{l} = ( l_1, l_2, l_3 )\transp$ and

\begin{align}
\mathtt{T}_i &= \mathbf{a}_i \mathbf{b}_4 \transp - \mathbf{a}_4 \mathbf{b}_i \transp \label{deftens} \text{,}
\end{align}

\noindent it can be verified that \eqnref{orig_constr} is equivalent to

\begin{align}
l_i = \mathbf{l}' \mathtt{T}_i \mathbf{l}'' \: \forall i \label{trifocal_line_constr} \text{.}
\end{align}

\begin{figure}[H]
\centering
\includegraphics[width=1.5in]{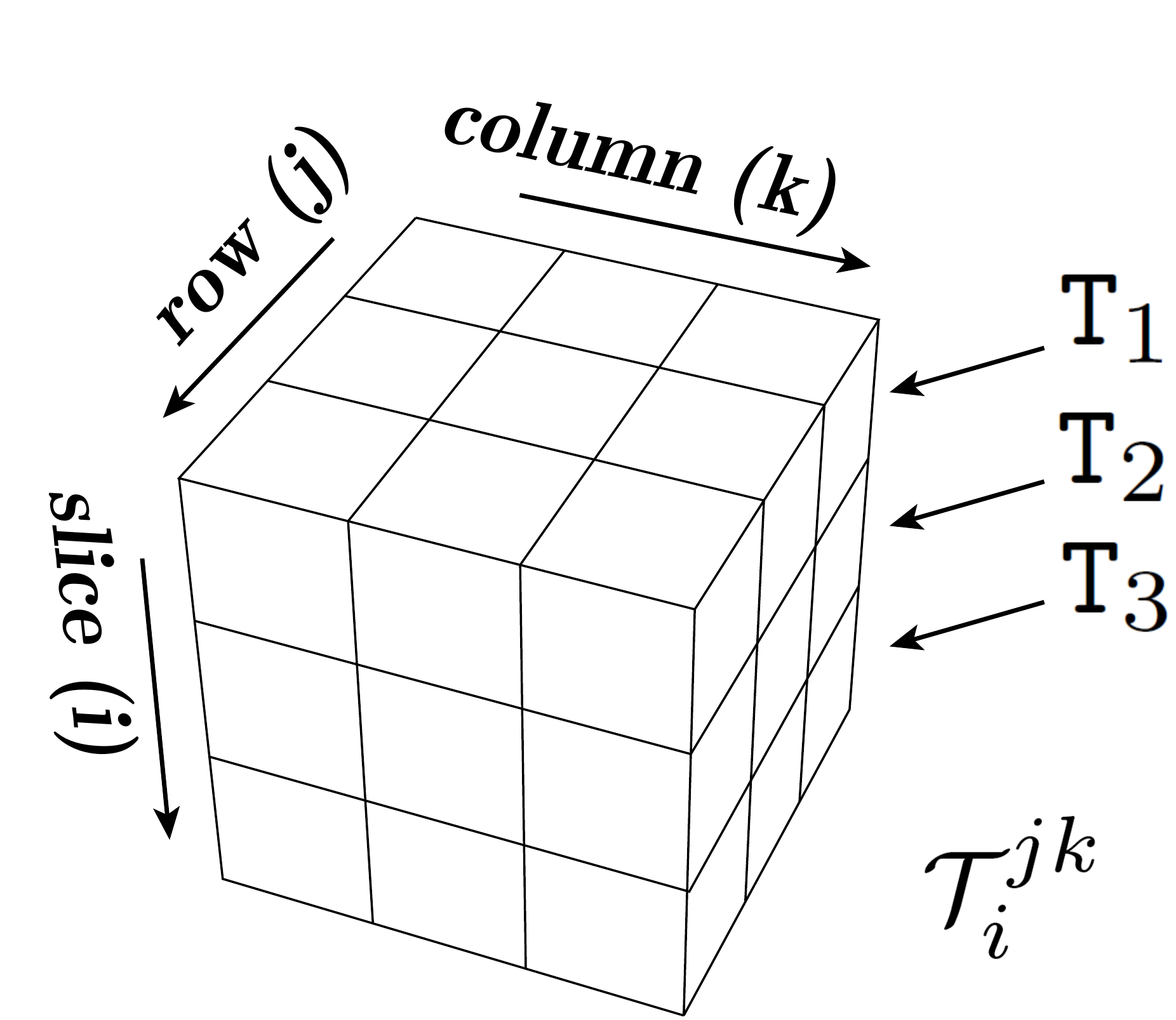}
\caption{Trifocal tensor parameters as a cube.}
\label{fig_trifocal_cube}
\end{figure}

Thus, the relationship between cameras has been completely described by $\{ \mathtt{T}_1, \mathtt{T}_2, \mathtt{T}_3\}$.  These three matrices, known as the \emph{correlation slices}, can be represented by a single $3 \times 3 \times 3$ tensor $\mathcal{T}$ (see \figref{fig_trifocal_cube}), allowing the above relations to be written equivalently in tensor notation as

\begin{align}
\mathcal{T}_i^{jk} = a_i^j b_4^k - a_4^j b_i^k\\
l_i = l'_j l''_k \mathcal{T}_j^{jk} \text{.}
\end{align}

It should be noted that, similarly to the fundamental matrix, the views are treated asymmetrically by the trifocal tensor.  In other words, there are three different trifocal tensors for any trio of views depending on the order that the views are considered in.  In the remainder of this work, we assume an implicit ordering of these views.

\section{Relationship to Camera Matrices} \label{part1}

Because the trifocal tensor provides a complete description of the epipolar geometry for three views, it must be possible to calculate a suitable set of camera matrices from the tensor.  However, it is not immediately obvious how one could factor a given tensor into the form of \eqnref{deftens} to get back the original camera matrices.  In this section, we explain how this may be accomplished, as described in \citep{HARTLEY2004}.

We begin by calculating the epipoles $\mathbf{e}'$ and $\mathbf{e}''$, which are the images of the focal point of the first camera in the other two views.  This is achieved in two steps.  First, denote the left and right null spaces of each $\mathtt{T}_i$ as $\mathbf{v}_i$ and $\mathbf{u}_i$ in

\begin{align}
\mathtt{T}_i \mathbf{v}_i &= \mathbf{0}, \quad i = 1 \ldots 3 \label{c1}\\
\mathtt{T}_i \transp \mathbf{u}_i &= \mathbf{0}, \quad i = 1 \ldots 3 \label{c2} \text{.}
\end{align}

Next, denote $\mathbf{U} = \left[ \mathbf{u}_1 | \mathbf{u}_2 | \mathbf{u}_3 \right]  \transp$ and $\mathbf{V} = \left[ \mathbf{v}_1 | \mathbf{v}_2 | \mathbf{v}_3 \right]  \transp$.  Then the epipoles (images of the first camera center in the other two images) are given by the null spaces of $\mathbf{U}$ and $\mathbf{V}$,

\begin{align}
\mathbf{U} \mathbf{e}' &= \mathbf{0} \label{c3}\\
\mathbf{V} \mathbf{e}'' &= \mathbf{0} \label{c4} \text{.}
\end{align}

Once the epipoles have been determined, one can recover the fundamental matrix between the first two views.  Recall that the tensor was defined based on a correspondence between lines $\mathbf{l} \leftrightarrow \mathbf{l}' \leftrightarrow \mathbf{l}''$ in each image.  If the third line $\mathbf{l}''$ back projects into a plane $\pi''$, then this plane induces a planar-homography mapping the first line $\mathbf{l}$ to the second line $\mathbf{l}'$.

A homography that transfers points according to $\mathbf{x}' = \mathbf{H x}$ transfers lines according to $\mathbf{l}' = \mathbf{H} \invtransp \mathbf{l}$.  According to this definition, \eqnref{trifocal_line_constr} implies that the homography transferring a line from the first to the second image induced by a line in the third image is given by

\begin{align}
\mathbf{H}_{12} = [\mathtt{T}_1, \mathtt{T}_2, \mathtt{T}_3 ] \mathbf{l}'' \text{,}
\end{align}

\noindent where the notational convention of writing $\mathbf{A} [ \mathbf{B}, \mathbf{C}, \mathbf{D} ] \mathbf{E}$ is used as a shorthand for $\left[ \mathbf{A} \mathbf{B} \mathbf{E} | \mathbf{A} \mathbf{C} \mathbf{E} | \mathbf{A} \mathbf{D} \mathbf{E} \right]$.

Given a point $\mathbf{x}$ in the first view, it is therefore transferred to $\mathbf{x}' = \mathbf{H}_{12} \mathbf{x}$ in the second view.  The line between two points is given by the cross product, so the epipolar line $\mathbf{l}_e'$ corresponding to $\mathbf{x}$ is given by

\begin{align}
\mathbf{l}_e' = \mathbf{e}' \times [\mathtt{T}_1, \mathtt{T}_2, \mathtt{T}_3 ] \mathbf{l}'' \mathbf{x} \text{.}
\end{align}

Thus, the fundamental matrix $\mathbf{F}_{12}$ from the first to the second view is given by

\begin{align}
\mathbf{F}_{12} &= [ \mathbf{e}' ]_\times [\mathtt{T}_1, \mathtt{T}_2, \mathtt{T}_3 ] \mathbf{l}'' \text{.}
\end{align}

This formula holds for any $\mathbf{l}''$ as long as $\mathbf{l}''$ is not in the null space of any $\mathbf{T}_i$.  One choice that avoids this degeneracy is $\mathbf{e}''$.  Thus, we obtain

\begin{align}
\mathbf{F}_{12} &= [ \mathbf{e}' ]_\times [\mathtt{T}_1, \mathtt{T}_2, \mathtt{T}_3 ] \mathbf{e}'' \text{.}
\end{align}

It is known that the fundamental matrix corresponding to a pair of cameras given by $\mathbf{P} = [ \identity | \mathbf{0}]$ and $\mathbf{P}' = [ \mathbf{M} | \mathbf{m} ]$ is equal to $[ \mathbf{m} ]_\times \mathbf{M}$.  Therefore, a suitable choice for the first two camera matrices consistent with the tensor are

\begin{align}
\mathbf{P} &= [\identity | \mathbf{0}] \label{c5}\\
\mathbf{P}' &= [ [\mathtt{T}_1, \mathtt{T}_2, \mathtt{T}_3 ] \mathbf{e}'' | \mathbf{e}' ] \label{c6} \text{.}
\end{align}

The third camera matrix can now be determined from \eqnref{deftens}.  Using the notation of \eqnref{formP3},

\begin{align}
\mathbf{a}_i &= \mathtt{T}_i \mathbf{e}'', \quad i = 1 \ldots 3 \\
\mathbf{a}_4 &= \mathbf{e}' \\
\mathbf{b}_4 &= \mathbf{e}''
\end{align}

and substituting into \eqnref{deftens} we obtain

\begin{align}
\mathtt{T}_i &= \mathtt{T}_i \mathbf{e}'' {\mathbf{e}''} \transp - \mathbf{e}' \mathbf{b}_i \transp \\
\mathbf{e}' \mathbf{b}_i \transp &= \mathtt{T}_i ( \mathbf{e}'' {\mathbf{e}''} \transp - \identity) \text{.}
\end{align}

If we choose the scale of $\mathbf{e}'$ such that ${\mathbf{e}'} \transp \mathbf{e}' = ||\mathbf{e}''|| = 1$, then we can left multiply by ${\mathbf{e}'} \transp$ to get

\begin{align}
\mathbf{b}_i \transp &= {\mathbf{e}'} \transp \mathtt{T}_i ( \mathbf{e}'' {\mathbf{e}''} \transp - \identity) \\
\mathbf{b}_i &= ( \mathbf{e}'' {\mathbf{e}''} \transp - \identity) \mathtt{T}_i \transp {\mathbf{e}'} \text{.}
\end{align}

Thus, a consistent choice for the third camera matrix is given by

\begin{align}
\mathbf{P}'' &= [ (\mathbf{e}'' {\mathbf{e}''} \transp - \identity ) [\mathtt{T}_1 \transp, \mathtt{T}_2 \transp, \mathtt{T}_3 \transp ] \mathbf{e}' | \mathbf{e}'' ] \label{c7} \text{.}
\end{align}

\section{Previously Known Constraints}

Because the tensor has 27 free parameters but the epipolar geometry that it represents has only 18 degrees of freedom, it is clear that there are dependencies between the tensor parameters.  These dependencies are the internal constraints, and one must verify that an estimated tensor satisfies all such constraints or else the tensor would not consistently encode for any meaningful geometry.

The derivation of constraints on the trifocal tensor has been a long and arduous process.  The first known constraints are the three rank constraints (3rd order) and two epipolar constraints (5th order) \citep{HARTLEY97Lines, FAUGERAS98Nonlinear, PAPADOPOULO98}, which are fairly straight-forward to identify from the definition.

The first set of sufficient constraints is due to \citep{FAUGERAS98Nonlinear}, where the 27 axes constraints were derived. These are 6th order polynomial constraints, but they are all independent from the existing rank and epipolar constraints.  Thus, a total of 32 constraints were needed at that time.

It was later shown in \citep{PAPADOPOULO98} that there are a set of 10 extended rank constraints (3 of which are equivalent to the original rank constraints).  These are 3rd order, and independent from the epipolar constraints.  Thus, together with the epipolar constraints, a total of only 12 constraints were needed.

Finally, \citep{CANTERAKIS00} showed that 8 new constraints (of up to 12th degree) could be derived based on the concept of generalized eigenvalues.  This is the only minimal set of necessary and sufficient constraints that is currently known.  However, a somewhat unsatisfactory property of these constraints is that they are not so simple as the well known rank and epipolar constraints.

\subsection{Rank and Epipolar Constraints}

The rank and epipolar constraints are a set of 5 independent constraints on the tensor that have been known for some time \citep{HARTLEY97Lines, FAUGERAS98Nonlinear, PAPADOPOULO98}.  It is not hard to spot these constraints from investigation of \eqnsref{c1}{c7}.  Specifically, in order for the null spaces in \eqnsref{c1}{c2} to exist, it is clear that each $\mathtt{T}_i$ must be of rank 2.  This may also be deduced from the fact that $\mathtt{T}_i$ was defined as the sum of two outer products in \eqnref{deftens}.  Thus, the first three constraints may be taken as,

\begin{align}
\rank \: \mathtt{T}_i &= 2, \quad i = 1 \ldots 3 \label{first_c} \text{.}
\end{align}

These are the \emph{rank constraints}.  They arise as a direct result of the assumption that, for corresponding lines, they must back-project into planes that intersect in a common 3D line.

Similarly, in order for the epipoles to exist as null spaces in \eqnsref{c3}{c4}, the rank of the matrix of left null vectors, as well as the rank of the matrix of right null vectors, must be 2.  Thus, an additional two constraints, called the \emph{epipolar constraints}, are given by

\begin{align}
\rank \: \mathbf{U} &= 2 \label{cu} \\
\rank \: \mathbf{V} &= 2 \label{cv} \text{.}
\end{align}

These simply enforce the property that all projection rays emanating from one camera's focal point must intersect at a common epipolar point in the image plane of another camera.  In other words, it enforces the equations of perspective projection.

Constraints of the form $\rank \mathbf{M}= 2$ on $3\times3$ matrices are usually treated as being equivalent to $\det \mathbf{M} = 0$ to be enforced as polynomials.  However, it should be noted that degenerate matrices having $\rank < 2$ do not represent valid trifocal tensors, although the inequality constraint that $\rank > 1$ does not affect the dof.

\subsection{Axes Constraints} \label{sec_axes_constr}

There are 27 axes constraints \citep{FAUGERAS98Nonlinear, FAUGERAS01Book}, 9 for each dimension of the tensor.  These are 6th degree polynomials in the tensor elements, independent from the rank and epipolar constraints.  Recall that the trifocal tensor elements are indexed as $\mathcal{T}_i^{jk}$ where $i,j,k$ index the correlation slice, row, and column dimensions, respectively.

In order to denote these constraints, we introduce the notation that an asterisk in any dimension means all the values in that dimension will be collected into a \emph{column} vector.  In other words, $\mathcal{T}_i^{*k}$ denotes the $k$th column of the $i$th correlation slice $\mathtt{T}_i$, $\mathcal{T}_i^{j*}$ denotes the \emph{transpose} of the $j$th row of $\mathtt{T}_i$, and $\mathcal{T}_*^{jk}$ denotes the column vector formed by taking the $(j,k)$th element from each correlation slice.

Without further ado, the \emph{vertical constraints} are

\begin{align}
\left| \mathcal{T}_*^{ik} , \mathcal{T}_*^{il} , \mathcal{T}_*^{jl} \right|
\left| \mathcal{T}_*^{ik} , \mathcal{T}_*^{jk} , \mathcal{T}_*^{jl} \right| &-\\
\left| \mathcal{T}_*^{jk} , \mathcal{T}_*^{il} , \mathcal{T}_*^{jl} \right|
\left| \mathcal{T}_*^{ik} , \mathcal{T}_*^{jk} , \mathcal{T}_*^{il} \right| &= 0 \text{,}
\end{align}

\noindent the \emph{horizontal column} constraints are

\begin{align}
\left| \mathcal{T}_i^{*k} , \mathcal{T}_i^{*l} , \mathcal{T}_j^{*l} \right|
\left| \mathcal{T}_i^{*k} , \mathcal{T}_j^{*k} , \mathcal{T}_j^{*l} \right| &-\\
\left| \mathcal{T}_j^{*k} , \mathcal{T}_i^{*l} , \mathcal{T}_j^{*l} \right|
\left| \mathcal{T}_i^{*k} , \mathcal{T}_j^{*k} , \mathcal{T}_i^{*l} \right| &= 0 \text{,}
\end{align}

\noindent and the \emph{horizontal row} constraints are

\begin{align}
\left| \mathcal{T}_i^{k*} , \mathcal{T}_i^{l*} , \mathcal{T}_j^{l*} \right|
\left| \mathcal{T}_i^{k*} , \mathcal{T}_j^{k*} , \mathcal{T}_j^{l*} \right| &-\\
\left| \mathcal{T}_j^{k*} , \mathcal{T}_i^{l*} , \mathcal{T}_j^{l*} \right|
\left| \mathcal{T}_i^{k*} , \mathcal{T}_j^{k*} , \mathcal{T}_i^{l*} \right| &= 0 \text{,}
\end{align}

\noindent where $i,j,k,l \in \{1,2,3\}$ and $i<j$ and $k<l$.  Note that commas have been used in the above equations to denote the merging of three columns together to form a $3 \times 3$ matrix.

\subsection{Extended Rank Constraints}

It is known that any linear combination of the tensor slices must also have rank 2.  Specifically, if $\mathbf{x} = ( x_1, x_2, x_3) \transp$ then

\begin{align}
\rank \sum_i x_i \mathtt{T}_i = 2 \text{.} \label{eqn_extended_rank}
\end{align}

Geometrically, if $\mathbf{x}$ is a point in the first image then the left and right null spaces of $\sum_i x_i \mathtt{T}_i$ are the corresponding epipolar lines in the second and third views, respectively \citep{HARTLEY2004}.

The requirement that all linear combinations have rank 2 can be translated into a set of 10 algebraic constraints called the \emph{extended rank constraints} \citep{PAPADOPOULO98, FAUGERAS01Book} as follows.  The rank constraint in \eqnref{eqn_extended_rank} implies a zero determinant, which can be expanded to the polynomial

\begin{align}
\begin{split}
\det \sum_i x_i \mathtt{T}_i &= c_1 x_1^3 + c_2 x_2^3 + c_3 x_3^3 \\[-10pt]
&+ c_4 x_1^2 x_2 + c_5 x_1^2 x_3\\
&+ c_6 x_2^2 x_1 + c_7 x_2^2 x_3\\
&+ c_8 x_3^2 x_1 + c_9 x_3^2 x_2\\
&+ c_{10} x_1 x_2 x_3 = 0 \text{.}
\end{split}
\end{align}

In order for this polynomial to be zero for \emph{all} choices of $\mathbf{x}$, it must be the case that the coefficients $c_i = 0 \: \forall i$.  These are the ten extended rank constraints.  The first three coefficients are $c_i = \det \mathtt{T}_i$ for $i = 1 \ldots 3$, so these are just the original rank constraints, but the remaining 7 are independent from the basic rank and epipolar constraints.  They may be expanded to

{
\fontsize{7}{8.4}\selectfont
\begin{align}
c_4 = \left| \mathcal{T}_1^{*1} , \mathcal{T}_1^{*2} , \mathcal{T}_2^{*3} \right|
+ \left| \mathcal{T}_1^{*1} , \mathcal{T}_2^{*2} , \mathcal{T}_1^{*3} \right|
+ \left| \mathcal{T}_2^{*1} , \mathcal{T}_1^{*2} , \mathcal{T}_1^{*3} \right| &= 0 \\
c_5 = \left| \mathcal{T}_1^{*1} , \mathcal{T}_1^{*2} , \mathcal{T}_3^{*3} \right|
+ \left| \mathcal{T}_1^{*1} , \mathcal{T}_3^{*2} , \mathcal{T}_1^{*3} \right|
+ \left| \mathcal{T}_3^{*1} , \mathcal{T}_1^{*2} , \mathcal{T}_1^{*3} \right| &= 0 \\
c_6 = \left| \mathcal{T}_2^{*1} , \mathcal{T}_2^{*2} , \mathcal{T}_1^{*3} \right|
+ \left| \mathcal{T}_2^{*1} , \mathcal{T}_1^{*2} , \mathcal{T}_2^{*3} \right|
+ \left| \mathcal{T}_1^{*1} , \mathcal{T}_2^{*2} , \mathcal{T}_2^{*3} \right| &= 0 \\
c_7 = \left| \mathcal{T}_2^{*1} , \mathcal{T}_2^{*2} , \mathcal{T}_3^{*3} \right|
+ \left| \mathcal{T}_2^{*1} , \mathcal{T}_3^{*2} , \mathcal{T}_2^{*3} \right|
+ \left| \mathcal{T}_3^{*1} , \mathcal{T}_2^{*2} , \mathcal{T}_2^{*3} \right| &= 0 \\
c_8 = \left| \mathcal{T}_3^{*1} , \mathcal{T}_3^{*2} , \mathcal{T}_1^{*3} \right|
+ \left| \mathcal{T}_3^{*1} , \mathcal{T}_1^{*2} , \mathcal{T}_3^{*3} \right|
+ \left| \mathcal{T}_1^{*1} , \mathcal{T}_3^{*2} , \mathcal{T}_3^{*3} \right| &= 0 \\
c_9 = \left| \mathcal{T}_3^{*1} , \mathcal{T}_3^{*2} , \mathcal{T}_2^{*3} \right|
+ \left| \mathcal{T}_3^{*1} , \mathcal{T}_2^{*2} , \mathcal{T}_3^{*3} \right|
+ \left| \mathcal{T}_2^{*1} , \mathcal{T}_3^{*2} , \mathcal{T}_3^{*3} \right| &= 0 \\
\begin{split}
c_{10} = \left| \mathcal{T}_1^{*1} , \mathcal{T}_2^{*2} , \mathcal{T}_3^{*3} \right|
+ \left| \mathcal{T}_1^{*1} , \mathcal{T}_3^{*2} , \mathcal{T}_2^{*3} \right|
+ \left| \mathcal{T}_2^{*1} , \mathcal{T}_1^{*2} , \mathcal{T}_3^{*3} \right| &+ \\
\left| \mathcal{T}_2^{*1} , \mathcal{T}_3^{*2} , \mathcal{T}_1^{*3} \right|
+ \left| \mathcal{T}_3^{*1} , \mathcal{T}_1^{*2} , \mathcal{T}_2^{*3} \right|
+ \left| \mathcal{T}_3^{*1} , \mathcal{T}_2^{*2} , \mathcal{T}_1^{*3} \right| &= 0 \text{.}
\end{split}
\end{align}
}

\subsection{Generalized Eigenspace Constraints}

It was shown in \citep{CANTERAKIS00} that a minimal set of 8 necessary and sufficient constraints for a trifocal tensor could be derived by ensuring that the generalized eigenspaces between each pair of tensor slices $\mathtt{T}_i$ intersect in a common point, and that there exists a common one-dimensional generalized eigenspace of all pairs.  Specifically, the following conditions must be satisfied:

\begin{enumerate}
  \item The polynomial $\det(\mathtt{T}_2 - \lambda \mathtt{T}_1)$ must have a single root $\lambda_1$ and double root $\lambda_2$, and $\rank(\mathtt{T}_2 - \lambda_2 \mathtt{T}_1) = 1$.
  \item The polynomial $\det(\mathtt{T}_3 - \mu \mathtt{T}_1)$ must have a single root $\mu_1$ and a double root $\mu_2$, and $\rank(\mathtt{T}_3 - \mu_2 \mathtt{T}_1) = 1$.
  \item If $\mathbf{a}, \mathbf{b}, \mathbf{a}', \mathbf{b}'$ are the generalized eigenvectors corresponding to the eigenvalues $\lambda_1$, $\lambda_2$, $\mu_1$ and $\mu_2$ (respectively), then $\mathbf{a} \propto \mathbf{a'}$.
\end{enumerate}

These constraints are not independent from any of the previous constraints.  The requirement that $\mathbf{a} \propto \mathbf{a'}$ amounts to checking for equality between two inhomogeneous 2-vectors and therefore provides two algebraic constraints.

In order for a 3rd degree polynomial

\begin{align}
p(\lambda) = a \lambda^3 + b \lambda^2 + c \lambda + d
\end{align}

\noindent to have a double root, it must be the case that

\begin{align}
B^2 - 4 AC = 0 \text{,}
\end{align}

\noindent where $A = b^2 - 3ac$, $B = bc - 9ad$ and $C = c^2 -3bd$.  In this case, the roots are given by

\begin{align}
\lambda_1 &= B/A - b/a \\
\lambda_2 &= - B/(2A) \text{.}
\end{align}

The two conditions of this type therefore provide two constraints.  In terms of the tensor elements, the coefficients of $\det(\mathtt{T}_2 - \lambda \mathtt{T}_1)$ are 3rd degree, $A,B,C$ are 6th degree, and the double root constraint is therefore 12th degree.

The constraint $\rank(\mathtt{T}_2 - \lambda_2 \mathtt{T}_1) = 1$ is equivalent to

\begin{align}
(\mathtt{T}_2 - \lambda_2 \mathtt{T}_1) (\mathbf{a} \times \mathbf{b}) \propto \mathtt{T}_1 \mathbf{a} \text{,} \end{align}

\noindent (and similarly for the second pair).  These requirements provide the final four constraints.

\section{New Circular Constraints} \label{sec_circular}

Clearly, there must exist a set of 3 constraints that can be supplanted with the well-known rank and epipolar constraints to yield a minimal set of necessary and sufficient constraints.  This set would likely be simpler and preferable for the purpose of constrained estimation than the generalized eigenspace constraints which are rather complicated.  However, the identification of those constraints has remained an open problem until now.

Specifically, we will show that the final 3 constraints can be obtained by substituting the recovered camera matrices from \eqnref{c6} and \eqnref{c7} back into \eqnref{deftens}.  This may at first seem like circular logic because the camera matrices were derived from \eqnref{deftens}.  However, the presence of singular outer-product matrices prevents the resulting equations from being simplified down to a trivial result such as $\identity = \identity$, and real constraints arise.  We will refer to these as the \emph{circular constraints}.

In \secref{part1} we have already derived

\begin{align}
\mathbf{a}_i &= \mathtt{T}_i \mathbf{e}'', \quad i = 1 \ldots 3 \\
\mathbf{a}_4 &= \mathbf{e}' \\
\mathbf{b}_i \transp &= {\mathbf{e}'} \transp \mathtt{T}_i ( \mathbf{e}'' {\mathbf{e}''} \transp - \identity), \quad i = 1 \ldots 3 \\
\mathbf{b}_4 &= \mathbf{e}'' \text{,}
\end{align}

\noindent which can be substituted back into \eqnref{deftens} to obtain

\begin{align}
\mathtt{T}_i &= \mathtt{T}_i \mathbf{e}'' {\mathbf{e}''} \transp - \mathbf{e}'( {\mathbf{e}'}\transp \mathtt{T}_i ( \mathbf{e}'' {\mathbf{e}''} \transp - \identity ) )\\
0 &= \mathtt{T}_i ( \mathbf{e}'' {\mathbf{e}''} \transp - \identity ) - \mathbf{e}' {\mathbf{e}'}\transp \mathtt{T}_i (\mathbf{e}'' {\mathbf{e}''} \transp - \identity )\\
0 &= (\identity - \mathbf{e}' {\mathbf{e}'}\transp) \mathtt{T}_i ( \mathbf{e}'' {\mathbf{e}''} \transp - \identity ) \text{.}
\end{align}

Because outer product matrices and $\mathtt{T}_i$ are all singular, nothing further can be canceled out.  However, one must be careful because this result was derived under the assumption that the epipoles were normalized, and only holds under that assumption.  Thus, it may be generalized to

\begin{align}
0 &= (||\mathbf{e}'||^2 \identity - \mathbf{e}' {\mathbf{e}'}\transp) \mathtt{T}_i ( \mathbf{e}'' {\mathbf{e}''} \transp - ||\mathbf{e}''||^2 \identity ) \label{all27} \text{.}
\end{align}

Because \eqnref{all27} is a $3 \times 3$ matrix equation which holds for each choice of $i = 1 \ldots 3$, it provides a total of 27 constraints that must be satisfied by $\mathcal{T}$ in order to remain consistent with our definitions.  We will first explain our findings pertaining to the independence of these constraints.

To begin with, we denote the individual constraint equations arising from the matrix equation \eqnref{all27} as $\mathcal{C}_i^{jk}, \: ijk \in \{1,2,3\}$ corresponding to the $(j,k)$th equality using $\mathtt{T}_i$.  Our first finding is that these constraints are not trivially satisfied, and that they are independent from the rank and epipolar constraints.  Secondly, constraints on $\mathtt{T}_i$ are independent from constraints on $\mathtt{T}_j$ for $i \neq j$.  Thus, from the counting argument, it can be inferred that if any constraint $\mathcal{C}_i^{jk}$ is satisfied, then $\mathcal{C}_i^{jk} \: \forall jk$ are satisfied.  These two findings are written formally as

\begin{align}
\left( \begin{array}{c}
    \det \: \mathbf{U} = 0 \: \wedge \\
    \det \: \mathbf{V} = 0 \: \wedge \\
    \det \: \mathtt{T}_i = 0 \: \forall i
\end{array}\right) \:&\nRightarrow\: \left(\mathcal{C}_i^{jk} = 0 \right) \quad \forall ijk \label{claim1} \\
\left(  \mathcal{C}_i^{jk} = 0 \right) \:&\Rightarrow\: \left(\mathcal{C}_i^{jk} = 0 \: \forall jk\right) \quad \forall i \label{claim2} \text{,}
\end{align}

\noindent and proven in \secref{sec_proof}.  Therefore, one choice of the final three independent constraints may be taken as $\mathcal{C}_i^{22} \: \forall i$.  Assuming the epipoles have been normalized, $\mathcal{C}_i^{22}$ expands to

\begin{equation}
\small
\begin{split}
{\mathbf{e}'_1} {\mathbf{e}'_2} {\mathbf{e}''_1} {\mathbf{e}''_2} \mathcal{T}_i^{11} &+
{\mathbf{e}'_1} {\mathbf{e}'_2} ({\mathbf{e}''_2}^2 - 1) \mathcal{T}_i^{12} +
{\mathbf{e}'_1} {\mathbf{e}'_2} {\mathbf{e}''_2} {\mathbf{e}''_3} \mathcal{T}_i^{13} +\\
{\mathbf{e}''_1} {\mathbf{e}''_2} ({\mathbf{e}'_2}^2 - 1) \mathcal{T}_i^{21} &+
(1 - {\mathbf{e}''_2}^2 )(1 - {\mathbf{e}'_2}^2) \mathcal{T}_i^{22} +
{\mathbf{e}''_2} {\mathbf{e}''_3} ({\mathbf{e}'_2}^2 - 1 ) \mathcal{T}_i^{23} +\\
{\mathbf{e}'_2} {\mathbf{e}'_3} {\mathbf{e}''_1} {\mathbf{e}''_2} \mathcal{T}_i^{31} &+
{\mathbf{e}'_2} {\mathbf{e}'_3} ({\mathbf{e}''_2}^2 - 1) \mathcal{T}_i^{32} +
{\mathbf{e}'_2} {\mathbf{e}'_3} {\mathbf{e}''_2} {\mathbf{e}''_3} \mathcal{T}_i^{33} = 0, \\ i \in \{1,2,3\} \text{,}\label{final3}
\end{split}
\end{equation}

\noindent where the elements of the first epipole are denoted by $\mathbf{e}' = ( {\mathbf{e}'_1}, {\mathbf{e}'_2}, {\mathbf{e}'_3} ) \transp$, and similarly for $\mathbf{e}''$.

\subsection{New Parameterization} \label{sec_parameterization}

Having identified the final circular constraints, it becomes possible to directly solve for the basis vectors of the components of a geometrically valid trifocal tensor, suggesting a rather elegant new parameterization for the tensor by simply using the coordinates in these bases.

Specifically, four parameters can be used to describe the inhomogeneous coordinates of the epipoles $\mathbf{e}'$ and $\mathbf{e}''$.  The epipolar constraint demands that the null spaces of $\mathbf{U}$ and $\mathbf{V}$ be the epipoles.  Thus, if the elements of $\mathbf{U}$ are arranged into the column vector $\mathbf{u}$, then it must satisfy

\begin{align}
\mathbf{C}_u \mathbf{u} = \mathbf{0}
\text{,}
\end{align}

\noindent where $\mathbf{C}_u$ is a $3 \times 9$ constraint matrix given by

\begin{align}
\mathbf{C}_u = \left[
  \begin{array}{ccc}
    {\mathbf{e}'}\transp & \mathbf{0} & \mathbf{0} \\
    \mathbf{0} & {\mathbf{e}'}\transp & \mathbf{0} \\
    \mathbf{0} & \mathbf{0} & {\mathbf{e}'}\transp \\
  \end{array}
\right] \text{.}
\end{align}

The space of all $\mathbf{u}$ satisfying this constraint is given by finding a basis $\mathbf{B}_u$ for the null space of $\mathbf{C}_u$ using Gauss-Jordan elimination.  Because there are 9 parameters and 3 constraints, there is a 6 dimensional basis for the null space.  One must be careful, however, to avoid the singular condition that arises if $\mathbf{U}$ has rank 1.  This can be enforced by ensuring that each coordinate in the basis is non-zero.  Also, the overall scale is irrelevant, so the last coordinate can be assumed equal to 1 and therefore $\mathbf{U}$ can be represented with 5 parameters (and similarly for $\mathbf{V}$).

The rank constraint demands that the left and right null spaces of each $\mathtt{T}_i$ are defined by $\mathbf{u}_i$ and $\mathbf{v}_i$.  This provides six constraints on each $\mathtt{T}_i$.  A seventh constraint is given by the circular constraint $\mathcal{C}_i^{22}$, and once again, the space of matrices satisfying these constraints is described by a basis for the null space of the constraint matrix.

To be precise, if the desired left and right null spaces of the slice $\mathtt{T}_i$ are $\mathbf{u}_i = (u_1, u_2, u_3)\transp$ and $\mathbf{v}_i = (v_1, v_2, v_3)\transp$, then the $7 \times 9$ constraint matrix is

\begin{align}
\mathbf{C}_t = \left[
  \begin{array}{ccccccccc}
    u_1 & 0 & 0 & u_2 & 0 & 0 & u_3 & 0 & 0 \\
    0 & u_1 & 0 & 0 & u_2 & 0 & 0 & u_3 & 0\\
    0 & 0 & u_1 & 0 & 0 & u_2 & 0 & 0 & u_3 \\
    v_1 & v_2 & v_3 & 0 & 0 & 0 & 0 & 0 & 0\\
    0 & 0 & 0 & v_1 & v_2 & v_3 & 0 & 0 & 0\\
    0 & 0 & 0 & 0 & 0 & 0 & v_1 & v_2 & v_3 \\
    a & b & c & d & e & f & g & h & i
  \end{array}
\right] \text{,}
\end{align}

\noindent where $a,b,c,\ldots,i$ are the coefficients of \eqnref{final3}.

Although there are 7 constraints, there is a linear dependency between the first 6 so the dimension of the null space is 3.  The first slice $\mathtt{T}_1$ can be represented with 2 parameters due to the overall scale ambiguity, but the remaining two slices require 3 parameters because they cannot be scaled independently. Thus, a total of 22 parameters are used to represent the tensor under this parameterization.

We now describe the reverse mapping for representing a given tensor in this parameterization.  First, the epipoles are computed from the null spaces as shown in \eqnsref{c1}{c4}.  If the tensor is not perfectly consistent, these null spaces may not exist so they should be extracted using the right singular vector, which provides a least squares estimate of a basis for the null space.

Once the epipoles have been found, the first four parameters are given by their inhomogeneous coordinates.  Then the constraint matrix $\mathbf{C}_u$ can be formed and the basis $\mathbf{B}_u$ can be found.  The coordinates $\mathbf{p}_u$ of $\mathbf{U}$ with respect to this basis may be found by solving a linear least squares system,

\begin{align}
\mathbf{B}_u \mathbf{p}_u &= \mathbf{u} \text{.}
\end{align}

After solving for $\mathbf{p}_u$ it should be normalized such that the last coordinate is equal to 1 in order to reduce the parameterization.  The same process can be used to obtain the coordinates for $\mathbf{V}$, as well as each slice $\mathtt{T}_i$ of the tensor.

\subsection{Proof} \label{sec_proof}

Our derivation of the circular constraints given in \eqnref{all27} shows that these 27 equalities must be true for internal consistency, although we have not yet proven our claims in \eqnsref{claim1}{claim2}.  In other words, we have not yet proven that these equalities are not implied by the rank and epipolar constraints, and that the subset in \eqnref{final3} are mutually independent.

If we assume that the circular constraints are dependent on the rank and/or epipolar constraints and then find a tensor that satisfies all the rank and epipolar constraints without satisfying the circular constraints, then we have reached a contradiction.  Thus, finding such a tensor would prove that the circular constraints are, in general, independent from the rank and epipolar constraints.

It is easy to find an unlimited number of counter-examples of this type by using our parameterization in \secref{sec_parameterization} to generate a random tensor that satisfies all constraints \emph{except} for the circular constraint by simply omitting the last row from $\mathbf{C}_t$.  It can then be verified that the circular constraints are not satisfied.  For example, we have used Maple with rational arithmetic to find the following tensor:

\begin{align}
\mathtt{T}_1 &= \left[
  \begin{array}{ccc}
357500/180469 & 200/251 & 475/251 \\
1500/719 & 0 & 3 \\
1700/719 & 2 & 1
  \end{array}
\right] \label{counterT1}\\
\mathtt{T}_2 &= \left[
  \begin{array}{ccc}
2050000/961197 & 200/401 & 1100/401 \\
8000/2397 & 1 & 4 \\
1500/799 & 0 & 3
  \end{array}
\right]\\
\mathtt{T}_3 &= \left[
  \begin{array}{ccc}
950000/480799 & 400/401 & 1100/401 \\
2500/1199 & 0 & 5 \\
4500/1199 & 4 & 1
  \end{array}
\right] \label{counterT3}\text{.}
\end{align}

It may be verified that the above tensor slices are rank 2 (rank constraints).  Extracting their null spaces and arranging them to form $\mathbf{U}$ and $\mathbf{V}$, we obtain

\begin{align}
\mathbf{U} &= \left[
  \begin{array}{ccc}
-251/100 & 5/4 & 1 \\
-401/100 & 2 & 1 \\
-401/100 & 2 & 1
  \end{array}
\right] \label{counterU}\\
\mathbf{V} &= \left[
  \begin{array}{ccc}
-719/500 & 6/5 & 1 \\
-799/500 & 4/3 & 1 \\
-1199/500 & 2 & 1
  \end{array} \right] \text{.} \label{counterV}
\end{align}

Again, these matrices are exactly rank 2 (epipolar constraints).  Extracting their null spaces yields the epipoles,

\begin{align}
\mathbf{e}' &= ( 100, 200, 1 ) \transp \\
\mathbf{e}'' &= ( -500, -600, 1 ) \transp \text{.}
\end{align}

Finally, we evaluate \eqnref{all27} and observe that none of the circular constraints are zero.  For brevity, we print only the values of the central constraints,

\begin{align}
\mathcal{C}_1^{22} &= -101022670792200/1834807869906823 \\
\mathcal{C}_2^{22} &=  -5236581973887/55211191885087 \\
\mathcal{C}_3^{22} &=  -14516209041800/698318420372419 \text{.}
\end{align}

Thus, the tensor in \eqnsref{counterT1}{counterT3} is not a valid trifocal tensor, and the circular constraints are indeed independent from the rank and epipolar constraints.

Proving that the three central constraints are independent from one another is trivial from the design of our algorithm for consistently parameterizing the tensor.  Specifically, one notices that once $\mathbf{U}$ and $\mathbf{V}$ have been calculated, there are no further dependencies between the $\mathtt{T}_i$ matrices.  Thus, it is not possible for $\mathcal{C}_i^{22}$ to have any dependency on $\mathcal{C}_j^{22}$ for $i \neq j$.

Having proven that $\mathcal{C}_i^{22} \: \forall i$ are mutually independent, and also independent from the rank and epipolar constraints, it can be seen from the counting argument that all degrees of freedom of the tensor have been accounted for.  This proves our claim in \eqnref{claim2}.  Indeed, if the circular constraint is added back into $\mathbf{C}_t$ in this example, then it can be verified that all of the constraints in \eqnref{all27} are satisfied exactly.

\subsection{Polynomial Form} \label{sec_polynomial}

The rank, extended rank, and axes constraints are all formulated as polynomials on the tensor elements already. The epipolar constraints can also be written as polynomials on the tensor elements by expressing the null vectors $\mathbf{u}_i$ and $\mathbf{v}_i$ in terms of the tensor elements.

A basis for the null space of any $n \times n$ matrix having rank $n-1$ can be computed in closed form by eliminating a row or column and taking $(n-1) \times (n-1)$ sub-determinants.  Once the null spaces are known, it is straight forward to plug these into the rule of Sarrus.  Because each element of $\mathbf{U}$ and $\mathbf{V}$ is quadratic in the tensor elements, and the determinant of a $3 \times 3$ is cubic in the elements, the epipolar constraints are therefore 5th degree.

However, it is often forgotten that it is necessary to eliminate a row or column that is linearly dependent on another row or column; otherwise, a zero vector will be extracted rather than a basis for the null space.  Replacing any $\mathbf{u}_i$ with a zero vector only means that $\det \mathbf{U} = 0$ will be trivially satisfied.  Thus, the particular polynomial that must be enforced by the 2 epipolar constraints must be chosen after identifying which rows are linearly independent.

Alternatively, the epipolar constraints could be translated into a larger set of equivalent polynomial constraints by considering all rank 2 possibilities of the $\mathtt{T}_i$ matrices.  In particular, since only two of the three rows or columns will be linearly independent, it would be necessary to consider two possibilities for each $\mathtt{T}_i$.  This leads to $2^3$ possible choices for each of $\mathbf{e}'$ and $\mathbf{e}''$, so the 2 epipolar constraints can be represented by 16 \emph{fixed} polynomial constraints.

The elements of $\mathbf{e}'$ and $\mathbf{e}''$ can be extracted in closed form using the same techniques, and their elements will be 4th degree polynomials because they are constructed from $2 \times 2$ determinants of a matrix having quadratic elements.  Examining \eqnref{final3}, the order of the circular constraints is therefore $4+4+4+4+1 = 17$.

The circular constraints are only trivially satisfied if \emph{both} $\mathbf{e}'$ and $\mathbf{e}''$ are zero vectors, so an incorrect choice about which rows or columns of $\mathtt{T}_i$ are linearly independent would make the constraints in \eqnref{final3} appear to be violated, when they might actually be satisfied.  Thus, it does not appear to be possible to enforce them as a set of fixed polynomials.  However, the extended rank or axes constraints could be used instead to make a sufficient set of fixed polynomials, if that were desired for some reason.

\section{Conclusions}

The primary contribution of this paper is an increased understanding of the trifocal tensor by deriving the 3 remaining constraints that can be combined with the well known rank and epipolar constraints to provide a minimal set of necessary and sufficient constraints.  This is a very satisfying discovery from a theoretical standpoint, and also has practical applications such as allowing for the new parameterization that we discuss.

To recapitulate, there are now four known sets of sufficient constraints that may be used to define a trifocal tensor, two of which are minimal.  In order of discovery, they are

\begin{enumerate}
  \item 3 rank + 2 epipolar + 27 axes
  \item 2 epipolar + 10 extended rank
  \item (minimal) 8 generalized eigenvalue
  \item (minimal) 3 rank + 2 epipolar + 3 circular.
\end{enumerate}

The newly found circular constraints were derived by a circular substituion; from the definition of the tensor, the general form for camera matrices was calculated and then substituted back into the definition of the tensor.  This resulted in constraint equations that were nontrivially satisfied only because of the existence of rank 1 outer product matrices that prevented the equation from being simplified down to $\identity = \identity$.

Thus, the new constraints do not particularly represent any new geometrical restrictions, but are simply another result of algebraic consistency required from the original constraint that corresponding lines must back-project into planes that intersect in a common line in 3D space.

\section*{Acknowledgements}

We thank Dr. Margaret J. Eppstein and Dr. Jan-Michael Frahm for their useful comments.  This work was sponsored in part by the Army Research Office under grant W911NF-09-1-0458.

\FloatBarrier
\bibliographystyle{elsarticle-num-names}
\bibliography{trifocal_tensor}

\end{document}